\documentclass{article}

\usepackage{microtype}
\usepackage{graphicx}
\usepackage{subfigure}
\usepackage{booktabs} 
\usepackage{amsmath} 

\usepackage{hyperref} 

\usepackage{amsmath,amsfonts,bm}

\def\eqref#1{equation~\ref{#1}}

\def\1{\bm{1}}

\def\mX{{\bm{X}}}

\DeclareMathAlphabet{\mathsfit}{\encodingdefault}{\sfdefault}{m}{sl}
\SetMathAlphabet{\mathsfit}{bold}{\encodingdefault}{\sfdefault}{bx}{n}

\def\gL{{\mathcal{L}}}

\def\gS{{\mathcal{S}}}

\DeclareMathOperator*{\argmin}{arg\,min}

\DeclareMathOperator{\sign}{sign}

\newcommand{\sbr}[1]{\left[#1\right]}
\newcommand{\cbr}[1]{\left\{#1\right\}}

\def\bb{\begin{equation}} \def\ee{\end{equation}}

\newcommand{\abs}[1]{\left|{#1}\right|}

\usepackage{hyperref}
\usepackage{url}
\usepackage{gensymb}

\newcommand{\norm}[1]{\left\|#1\right\|}

\usepackage[utf8]{inputenc} 
\usepackage[T1]{fontenc}    
\usepackage{hyperref}       
\usepackage{url}            
\usepackage{booktabs}       
\usepackage{amsfonts}       
\usepackage{nicefrac}       
\usepackage{microtype}      
\usepackage{thm-restate}

\usepackage{xcolor}         

\usepackage{microtype}
\usepackage{graphicx}
\usepackage{subfigure}
\usepackage{changes}

\usepackage{multirow}
\usepackage{hhline}
\usepackage{wrapfig}
\usepackage{floatrow}
\usepackage{caption}
\usepackage{enumitem}
\usepackage[export]{adjustbox}
\usepackage{amsmath,mathtools}

\usepackage{algorithmic}
\usepackage[vlined,ruled]{algorithm2e}
\newcommand*{\Scale}[2][4]{\scalebox{#1}{$#2$}}%

\makeatletter
\newcommand{\raisemath}[1]{\mathpalette{\raisem@th{#1}}}
\newcommand{\raisem@th}[3]{\raisebox{#1}{$#2#3$}}
\makeatother

\newcommand{\uglad}{{\texttt{uGLAD}~}}
\newcommand{\ugladns}{{\texttt{uGLAD}}}
\newcommand{\glad}{{\texttt{GLAD}~}}

\newcommand{\Rho}{\mathrm{P}}

    \PassOptionsToPackage{numbers, compress}{natbib}

    \usepackage[preprint]{neurips_2022}

\title{\ugladns: Sparse graph recovery by optimizing deep unrolled networks}

\author{
  Harsh Shrivastava$^1$~~~~
  Urszula Chajewska$^1$~~~~
  Robin Abraham$^1$~~~~
  Xinshi Chen$^2$ \\
  \hspace{0mm}\\
  \hspace{-3mm}
  \begin{tabular}{c}
      $\prescript{1}{}{\text{Microsoft Research, Redmond, USA}}, ~~~\prescript{2}{}{\text{School of Mathematics, Georgia Institute of Technology}}$
  \end{tabular}
}

\begin{document}

\maketitle

\begin{abstract}

Probabilistic Graphical Models (PGMs) are generative models of complex systems. They rely on conditional independence assumptions between variables to learn sparse representations which can be visualized in a form of a graph. Such models are used for domain exploration and structure discovery in poorly understood domains. 
This work introduces a novel technique to perform sparse graph recovery by optimizing deep unrolled networks. Assuming that the input data $X\in\mathbb{R}^{M\times D}$ comes from an underlying multivariate Gaussian distribution, we apply a deep model on $X$ that outputs the precision matrix $\hat{\Theta}$,
which can also be interpreted as the adjacency matrix. 
Our model, \ugladns
\footnote{\small Code: \url{https://github.com/Harshs27/uGLAD}}
, builds upon and extends the state-of-the-art model \glad \cite{shrivastava2020glad} to the unsupervised setting.
The key benefits of our model are 
(1) \uglad automatically optimizes sparsity-related regularization parameters leading to better performance than existing algorithms. (2) We introduce multi-task learning based `consensus' strategy for robust handling of missing data in an unsupervised setting. 
We evaluate model results on synthetic Gaussian data, non-Gaussian data generated from Gene Regulatory Networks, and present a case study in anaerobic digestion.


\end{abstract}

\section{Introduction}

Probabilistic graphical models (PGMs) \cite{Pearl88, Koller2009ProbabilisticGM} are generative models of complex systems, used to describe dependencies within a set of random variables and visualize the structure of a domain.  The models rely on conditional independence assumptions between variables, which result in sparse representation and enable efficient inference.  In the graphical representation of the models, conditional independence is indicated by an absence of an edge between two variables.  Such models can be learned from observational data \cite{heckerman1995a, friedman2008sparse}.  Structure discovery enabled by PGMs is important for new and poorly understood domains where relationships between variables are not known. PGMs have been used in various domains, including medical diagnosis~\cite{heckerman1992toward1, heckerman1992toward2}, fault diagnosis, analysis of genomic data via gene regulatory networks~\cite{moerman2019grnboost2,pratapa2020benchmarking,aluru2021engrain,shrivastava2022grnular}, speech recognition, and financial applications~\cite{hallac2017network}.

The problem of recovering the structure from observational data is particularly difficult in high-dimensional settings, where the number of features may be larger than the number of observations. We focus specifically on learning undirected models where the data is assumed to generated from a multivariate Gaussian distribution~\cite{friedman2008sparse,belilovsky2017learning,zheng2018dags,yu2019dag,shrivastava2020glad}.  In such cases, the goal is to estimate a sparse inverse covariance matrix.  Sparsity constraint is typically enforced by the use of $\ell_1$ (lasso) regularization.

\begin{wrapfigure}[25]{R}{0.23\textwidth}
\centering
\includegraphics[width=25mm, height=83mm]{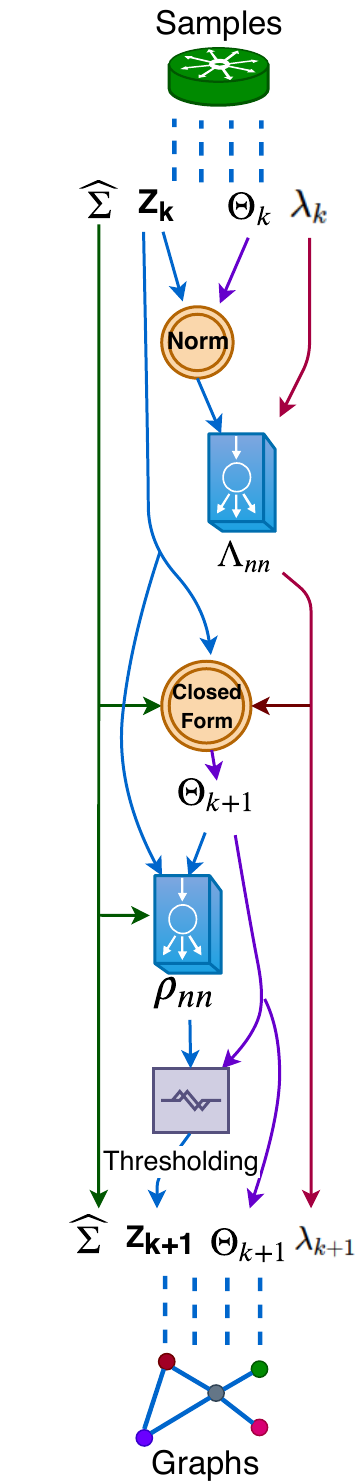}
\caption{\small The recurrent unit \texttt{GLADcell}. (Taken from~\cite{shrivastava2020glad})} 
\label{fig:glad-architecture}
\end{wrapfigure}

Assume we have a $d$-dimensional multivariate Gaussian random variable $X=[X_1,\ldots,X_d]^\top$ with $m$ observations. The goal is to estimate its covariance matrix $\Sigma^*$ and precision matrix $\Theta^* = (\Sigma^*)^{-1}$. 
$\Theta^*$ encodes conditional independence assumptions between variables:  the $ij$-th component is zero if and only if $X_i$ and $X_j$ are conditionally independent given the other variables $\cbr{X_k}_{\raisemath{1.5pt}{k\neq i,j}}$. 
This problem is known as the sparse graph recovery problem and usually formulated (following~\cite{friedman2008sparse}), as the $\ell_1$-regularized maximum likelihood estimation
\begin{align}
    \label{eq:sparse_concentration}
\widehat{\Theta} =  \argmin\nolimits_{\Theta \in \gS_{++}^d} - \log(\det{\Theta}) +  \text{tr}(\widehat{\Sigma}\Theta) + \rho\norm{\Theta}_{1,\text{off}},
\vspace{-1mm}
\end{align}
where $\widehat{\Sigma}$ is the empirical covariance matrix based on $m$ samples, $\gS_{++}^d$ is the space of $d\times d$ symmetric positive definite matrices (SPD), and $\norm{\Theta}_{1,\text{off}}=\sum_{i\neq j}|\Theta_{ij}|$ is the off-diagonal $\ell_1$ regularizer with regularization parameter $\rho$. The use of this estimator is justified even for non-Gaussian $X$, since it is minimizing an $\ell_1$-penalized log-determinant Bregman divergence~\cite{ravikumar2011high}. The problem in Eq.~\ref{eq:sparse_concentration} is a convex optimization problem.  It can be solved by many algorithms, see Sec.~\ref{sec:related-works} for examples.


However, classic approaches have their limitations in both the statistical aspect and computational aspect. Statistically, the formulation uses a single regularization parameter $\rho$ for all entries in the precision matrix $\Theta$, which may not be optimal.  A recent theoretical work~\cite{sun2018graphical} validates the use of adaptive parameters.  \cite{shrivastava2020glad}  has proposed a model with multiple regularization parameters called \glad and has shown emprical evidence that such a model pushes the sample complexity limits.
Based on that evidence, we hypothesise that one may obtain better recovery results by allowing the regularization parameters to vary across the entries in the precision matrix. However, it is hard for traditional approaches to search over a large number of hyperparameters. Computationally, the complexity of solving the optimization depends on the convexity of the objective, the step sizes of the algorithm, the initialization, the design of the update steps, etc. Different problems may require different designs of the algorithm to achieve a better efficiency. A unified design for all problems may not be optimal. 

In this work, we propose \uglad (unsupervised-\texttt{GLAD}), which is a deep model that can recover sparse graphs in an unsupervised manner.  As the name suggests, it builds upon and extends the \glad model, which recovers sparse graphs under supervision. 
\uglad uses the same objective function as \glad.
Using an additional variable $Z$, the $\ell_1$-regularized maximum likelihood from Equation \ref{eq:sparse_concentration} can be re-written as 
\begin{align}
\widehat{\Theta} = & \argmin\nolimits_{\Theta \in \gS_{++}^d} - \log(\det{\Theta}) +  \text{tr}(\widehat{\Sigma}\Theta) + \rho\norm{Z}_1,\\ \nonumber
&s.t.\quad Z=\Theta \nonumber
\vspace{-1mm}
\end{align}
Now, including the constraint as squared penalty term $\lambda$ we obtain the reformulated objective as 
\begin{equation}
\label{eq:uglad-objective}
\begin{aligned}
 \widehat{\Theta}_{\lambda}, \widehat{Z}_{\lambda} = &{\textstyle\argmin_{\Theta, Z \in \mathcal{S}_{++}^{d}}} - \log(\det{\Theta}) + \text{tr}(\widehat{\Sigma}\Theta) \\
& + \rho \norm{Z}_1 + {\textstyle \frac{1}{2}\lambda}\norm{Z-\Theta}^2_F
\end{aligned}
\vspace{-1mm}
\end{equation}
Note that introducing the variable $Z$ helps in splitting the objective into 2 parts and those can be optimized alternately using the Alternating Minimization algorithm.

Key contributions of this work:
\begin{itemize}[leftmargin=*,nolistsep]
    \item \textit{Extending \texttt{GLAD} to unsupervised setting}: The \uglad doesn't rely on availability of ground truth to do graph recovery. 
    \item \textit{Adaptive hyperparameters}: The \uglad architecture design enables the hyperparameters to optimally adapt at each step of the unrolled Alternating Minimization (AM) algorithm~\cite{shrivastava2020glad} that leads to its superior performance.
    \item \textit{Automatically decide optimum sparsity parameters}: The sparsity of the recovered graph is highly sensitive to the choice of the regularization hyperparameters. Instead, \uglad models 
    hyperparameters within the neural network framework 
    and they are directly optimized for the \uglad objective defined above. So, there is no need to 
    separately optimize 
    the sparsity hyperparameters which is otherwise a computationally expensive process.
    \item \textit{Runtime efficiency}: The \uglad software can run on GPUs for higher time efficiency and scalability. 
    \item \textit{Missing data handling}: The \uglad framework can also be used for multi-task learning. We leverage this property further to develop a novel `consensus' strategy to robustly handle missing data. 
\end{itemize}

\section{Related work}\label{sec:related-works}

\textbf{Traditional algorithms for graphical lasso}: These are primarily iterative methods for optimizing the graphical lasso objective. 
Main methods developed for this problem in the last two decades are detailed in this survey paper~\cite{witten2011new}. The variant of the Block Coordinate Descent (BCD) algorithm by~\cite{friedman2008sparse} is widely used to recover graphs using the graphical lasso method. This method is also implemented in the popular python `sklearn' package. The G-ISTA algorithm by~\cite{guillot2012iterative} is based on the iterative shrinkage thresholding procedure and one of the prominent methods based on using the proximal gradient descent approach. The Alternating Direction Method of Multipliers (ADMM)~\cite{danaher2014joint} 
has also been successfully used in various graphical lasso based applications.

\begin{wrapfigure}[30]{R}{0.4\textwidth}
  \centering
\includegraphics[width=50mm]{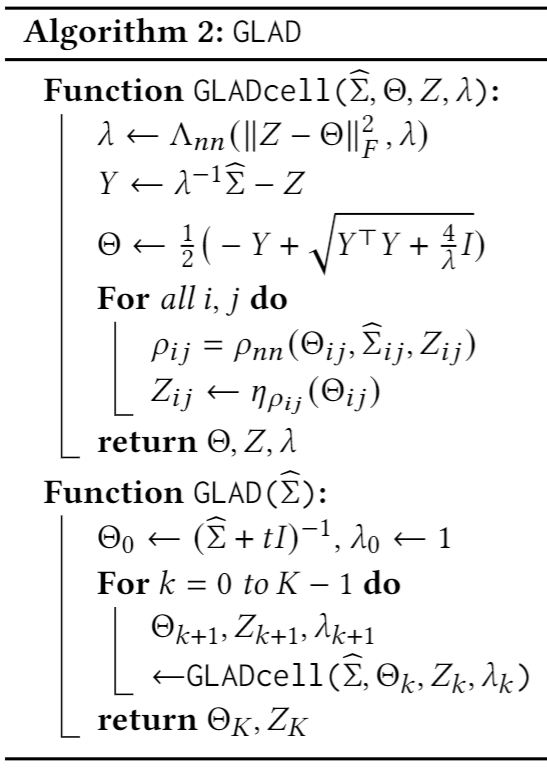}\\
\includegraphics[width=50mm, height=20mm]{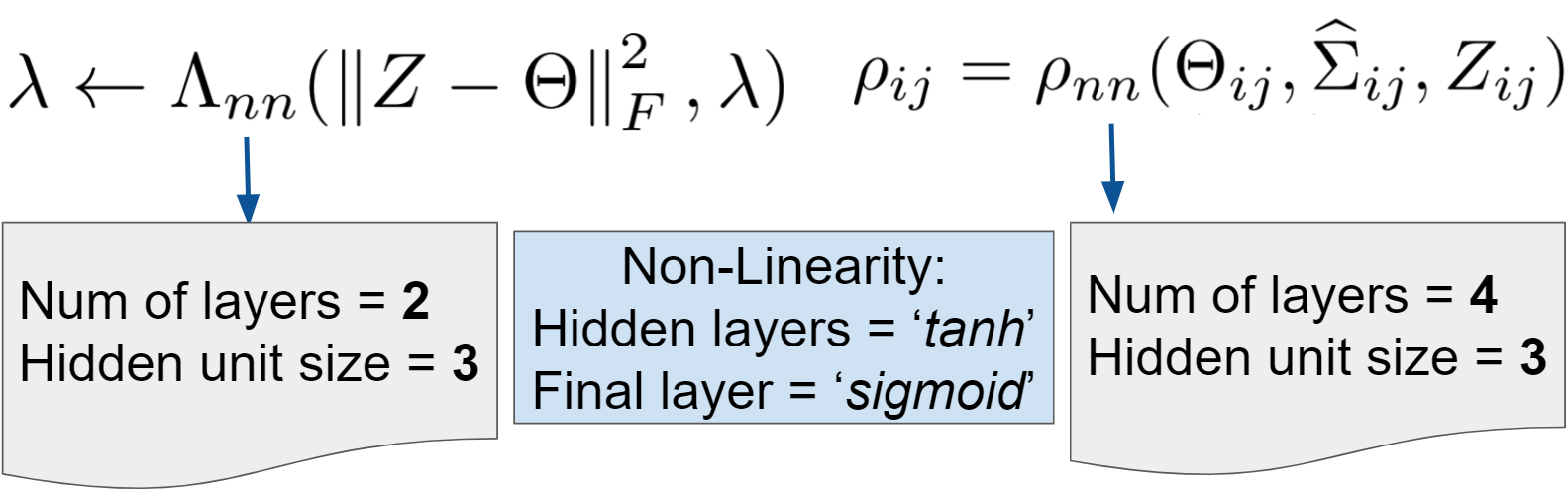}
\caption{\small {\bf Minimalist} neural network architectures designed for \texttt{GLAD} along with the optimizing algorithm.}
\label{fig:nn-design}
\end{wrapfigure}

\textbf{Deep Learning approaches for graph recovery}: DeepGraph (DG)~\cite{belilovsky2017learning}, is a supervised deep learning method that takes in the input samples and outputs the corresponding adjacency matrix which shows the connections between input features. DeepGraph architecture consists of many convolutional layers followed by multi-layer perceptrons that finally decides whether an edge is present between every combination of features. Another relevant deep learning method roughly based on modeling the input data with a Variational Autoencoder for graph recovery is DAG-GNN~\cite{yu2019dag}. These deep architectures have very high number of learnable parameters, which is a significant drawback. Hence, we pursue a different line of research (using inductive biases) which gives similar performance with significantly reduced number of learnable parameters and brings more interpretability as shown in \cite{shrivastava2020glad}.

\textbf{Deep learning models using inductive biases}: 
Improved performance often results from including domain knowledge in the design/initialization of deep learning architectures.
For instance,~\cite{shrivastava2019cooperative} presents a generic technique to use a probabilistic graphical model as a prior to design a deep model. The authors were able to show enhanced performance on the document classification task by leveraging the Latent Dirichlet Allocation prior. Another way of including prior knowledge about the domain is using an optimization algorithm for a related objective function as a template to design the deep architecture. Unrolling the optimization algorithms and parameterizing the step updates using neural networks have been fairly successful for 
many tasks~\cite{liu2019alista,chen2020rna,shrivastava2020using,pu2021learning,shrivastava2022grnular}.

This work focuses on recovering undirected graphs, specifically based on optimizing the graphical lasso objective function. Hence, we skipped discussing the methods developed specifically to recover Directed Acyclic Graphs (DAGs).

The work most closely related to ours is the \texttt{GLAD}~\cite{shrivastava2020glad} model. Since our algorithm builds upon \texttt{GLAD}'s architecture we are going to describe it in detail in Section \ref{sec:uglad}, while pointing out ways in which \uglad differs from \texttt{GLAD}.

\section{The \uglad model}
\label{sec:uglad}

Given input data $X\in\mathbb{R}^{M\times D}$, with $M$ samples 
with $D$ features, the task is to recover a sparse graph showing correlations between the $D$ features. We assume that the data comes from an underlying multivariate Gaussian distribution. Recovering the sparse graph (the adjacency matrix) corresponds to obtaining an estimate $\hat{\Theta}$ of  the precision matrix $\Theta^*$ of the Gaussian distribution.

\subsection{Understanding the \texttt{GLAD} architecture}

\uglad uses the same architecture as \glad. 
We we have a function $\Theta=f_{nn}(X)$, implemented as the \texttt{GLAD} architecture ~\cite{shrivastava2020glad}. The \texttt{GLAD} model uses the Alternating Minimization algorithm updates, unrolled to some iterations, for the maximum likelihood objective as a template for its deep architecture design. 

Penalty constants $(\rho,\lambda)$ are replaced by problem dependent neural networks, $\rho_{nn}$ and $\Lambda_{nn}$. These neural networks are minimalist in terms of the number of parameters as the input dimensions are mere $\{3, 2\}$ for $\{\rho_{nn}, \Lambda_{nn}\}$ and outputs a single value. Algorithm presented in Fig.~\ref{fig:nn-design} summarizes the update equations for the unrolled AM based model, \texttt{GLAD}. This unrolled algorithm with neural network augmentation can be viewed as a highly structured recurrent architecture as illustrated in Figure~\ref{fig:glad-architecture}.

The key features making the \texttt{GLAD} model a 
suitable choice are:
\begin{itemize}[leftmargin=*,nolistsep]
    \item Minimalist learnable parameters: The number of learnable parameters is significantly smaller than in other deep learning methods.
    \item Permutation invariance: The design of \texttt{GLAD} inherently maintains the permutation invariance w.r.t. the input.
    \item Positive definite constraint: The design prior of the AM algorithm enforces the positive definite constraint of the precision matrix at every step of its optimization. 
    \item Interpretability: One can see the state of the recovered graph at each unrolled step of the AM optimization, thus giving insights about the learning process. 
    \item Linear convergence: The AM algorithm converges linearly for the graphical lasso objective and therefore gives faster convergence rates.
\end{itemize}


For the sake of completeness of understanding the \texttt{GLAD} architecture, we graciously borrow the algorithm (see, Alg.2) 
 and neural network design details from~\cite{shrivastava2020glad} here.


\subsection{The \glad loss function}\label{sec:glad-loss}
To learn the parameters in \texttt{GLAD} architecture, the authors  used supervision in form of the true underlying graphs. They leveraged the interpretable nature of the \texttt{GLAD}'s deep architecture to define the loss for training. Specifically, each iteration of the model will output a valid precision matrix estimation and this allowed them to add auxiliary losses to regularize the intermediate results of \texttt{GLAD}, guiding it to learn parameters which can generate a smooth solution trajectory. 

The authors used Frobenius norm in their loss function:  
\begin{align}\label{eq:loss-glad}
    \gL_\text{\glad} := \frac{1}{n}\sum_{i = 1}^n \sum_{k=1}^K \gamma^{K-k} \norm{\Theta_{k}^{(i)} -\Theta^{*}}^2_F,
\end{align}
where $\Scale[0.92]{(\Theta_{k}^{(i)}, Z_{k}^{(i)}, \lambda_{k}^{(i)})=\texttt{GLADcell}_f(\widehat{\Sigma}^{(i)}, \Theta_{k-1}^{(i)}, Z_{k-1}^{(i)}, \lambda_{k-1}^{(i)})}$ is the output of the recurrent unit $\texttt{GLADcell}$ at $k$-th iteration, $K$ is number of unrolled iterations, $\gamma\leq 1$ is a discounting factor and $\Theta^*$ is the ground truth precision matrix. Then the stochastic gradient descent algorithm was used to train the parameters $f$ in the \texttt{GLADcell}.

\subsection{The \uglad loss function}\label{sec:uglad-loss}

In contrast to \glad, \uglad is designed to work without supervision, so it cannot take advantage of the underlying true generating distribution in the form of the true covariance matrix. Thus, we need to replace \texttt{GLAD}'s MSE loss from Equation~\ref{eq:loss-glad} with a new loss.  Given a function $\Theta=f_{nn}(X)$,  we optimize the log likelihood function given by
\begin{flalign}\label{eq:loss-uglad}
    \gL_\text{\ugladns}(S, \Theta) = -\log\abs{\Theta} + \langle S, \Theta\rangle
\end{flalign}
\begin{flalign}
    \gL_\text{\ugladns}(X) = -\log\abs{f_{nn}(X)} 
    + \langle\operatorname{cov}(X), f_{nn}(X)\rangle
\end{flalign}
where $S=\operatorname{cov}(X)$ is the covariance matrix. 
We take the function $f_{nn}$ as the \texttt{GLAD} model and substitute it in the \uglad loss function.




\subsection{\uglad: Key Features and Training}
The \texttt{GLAD} model was developed as a supervised learning model, where the supervision is provided in terms of pairs of input data $X$ and the ground truth precision matrices $\Theta^*$. The parameters were trained by optimizing the mean square error between the predicted and the true precision matrices, refer Sec.~\ref{sec:glad-loss}. In our model \uglad (Unsupervised \texttt{GLAD}), we extend the \texttt{GLAD} model to  the setting where the ground truth is unavailable. 
Note that this is a much more realistic scenario - in most real-world domains the data generating distribution is not known.
We replace the MSE loss with the \uglad loss described in Sec.~\ref{sec:uglad-loss}. Algorithm~\ref{algo-uGLAD} gives the pseudo code for learning the \uglad model for doing sparse graph recovery. 

\begin{wrapfigure}[10]{R}{0.5\textwidth}
\begin{minipage}{\textwidth}
\vspace{-6mm}
\begin{algorithm}[H]
\caption{Optimizing \uglad}
\label{algo-uGLAD}
\begin{algorithmic}
    \STATE Input: Observations $\mX\in\mathbb{R}^{M\times D}$
    \STATE $S = \operatorname{cov}(\mX)$\;
    \FOR{$e=1,\cdots,$ E}
        \STATE $\hat{\Theta}_e=\texttt{GLAD}(S)$ unrolled for L iterations\;
        \STATE Compute loss $\gL_{\ugladns}(S, \hat{\Theta}_e)$\;
        \STATE Backprop to update \texttt{GLAD} parameters\;
    \ENDFOR
    \STATE return $\hat{\Theta}_E$
\end{algorithmic}
\end{algorithm}
\end{minipage}
\end{wrapfigure}

\subsection{Convergence properties of \glad and \uglad}
The \glad paper \cite{shrivastava2020glad} evaluates convergence properties for the \glad algorithm using normalized mean square
error (NMSE) and probability of success (PS) to evaluate the algorithm performance. NMSE is 
$\log_{10}(\mathbb{E}\norm{\Theta^p-\Theta^*}_F^2/\mathbb{E}\norm{\Theta^*}^2_F)$  and PS is the probability of correct signed edge-set recovery, i.e., $\mathbb{P}\sbr{\sign(\Theta^p_{ij}) = \sign(\Theta^*_{ij}),\forall(i, j)\in\mathbf{E}(\Theta^*)}$, where $\mathbf{E}(\Theta^*)$ is the true edge set.

Optimization objective always converges.  However, errors of recovering true precision matrices measured by NMSE have very different behaviors given different regularity parameter $\rho$, which indicates the necessity of directly optimizing NMSE and hyperparameter tuning.
NMSE values are very sensitive to both $\rho$ and the quadratic penalty $\lambda$ of ADMM method. In \glad and \ugladns, $\rho$ and $\lambda$ are not fixed, but are optimized together with the rest of network parameters, leading to smooth convergence in NMSE.

In experiments evaluating edge recovery success, \glad 
consistently outperforms traditional methods in terms of sample complexity as it recovers the true
edges with considerably fewer number of samples.

Since, in \uglad we are still using the AM minimization based \glad architecture which is also based on optimizing the Eq.~\ref{eq:sparse_concentration}, we expect the linear convergence properties of the AM algorithm will hold for \uglad as well. The synthetic experiments in Sec.~\ref{sec:expts} show the results obtained from \uglad are better or even surpass in comparison to block coordinate descent based approach.

\section{Multi-task learning for precision matrix recovery}
\label{sec:multi-task}

Most of the work in learning Gaussian graphical models has focused on estimating a single model.  In recent years, the framework was extended to jointly fitting a collection of such models, based on data that share the same variables, with dependency structure varying with some external category.  For eg., in an NLP application, we can encounter different styles, which induce different links between some concepts, even as the underlying grammar and semantics of the language stay the same.

There have been extensive studies on the joint estimation of multiple undirected Gaussian graphical models~\cite{song2009keller,honorio2010multi,guo2011joint,cai2011constrained,oyen2012leveraging,danaher2014joint,kolar2010estimating,mohan2014node,peterson2015bayesian,yang2015fused,gonccalves2016multi,varici2021learning}. Most traditional algorithms construct a joint objective for multiple estimation tasks. This objective typically incorporates  similarities among various tasks by adding group norms or regularization terms. However, in many practical problems, we only know that multiple tasks are related, without knowing how they are similar to each other quantitatively. Manually constructing the joint objective may not best reflect the actual similarity.

In contrast to traditional algorithms, in \texttt{uGLAD}, we do not need to pre-assume the specific similarity among different tasks. Instead, we use a single network \uglad to solve multiple tasks. Since the parameters in \uglad are shared across different tasks, the similarity among the tasks is automatically learned from data.
More specifically, given samples from $K$ different models, $\mathbf{X_K} = [X_1, X_2, \cdots, X_K]$, we optimize the following objective


\begin{align}\label{eq:loss-mt-1}
    \gL_\text{\ugladns-multitask}&(\mathbf{X_K}) =\frac{1}{K}\sum_{k=1}^{K}\gL_\text{\ugladns}(\operatorname{cov}(X_k), f_{nn}(X_k))
\end{align}
Alternatively, we could use the following version with in-task data-split:
\begin{align}
    \gL_\text{\ugladns-multitask}(\mathbf{X_K}) =\frac{1}{K}\sum_{k=1}^{K}\gL_\text{\ugladns}(\operatorname{cov}(X_k^{val}), f_{nn}(X_k^{train}))
\end{align}

\section{Handling missing values}\label{sec:missing-data}
The missing data problem is ubiquitous in all data problems.  
\uglad can easily be extended to handle this problem.
First, 
let's examine different ways one can encounter missing values in our problem setting. For example, consider the input data $X\in \mathbb{R}^{M\times D}$ coming from the gene expression data. Here, $M$ (rows/samples) will be the samples collected from the microarray experiments and $D$ (columns/features) will be the different genes. The output graph will be the gene regulatory network showing how the genes are correlated to each other. Technical glitches can occur which can lead to erroneous recording or absence of expression values of certain genes leading to missing values in features. Similarly, some samples might be recorded incorrectly or face dropouts issue leading to missing values in samples. We discuss how to handle such cases of missing values in this section.

\textbf{Missing values in features:}\label{sec:missing-data-features}
If we observe that some specific feature columns have missing values and we have reasons to believe that these values are missing at random, we can run imputation algorithms for those columns to predict the missing entries. Then, we will have the complete imputed input data $X_{imp}\in \mathbb{R}^{M\times D}$ over which we can run the \uglad model and obtain the underlying precision matrix. Few of the imputation techniques that can be potentially utilized for this task are:
\begin{itemize}[leftmargin=*,nolistsep]
    \item \textit{Statistical imputation}: Simple statistical techniques like replacing the missing values in the column by its mean, median or a constant. A detailed analysis of similar techniques can be found in~\cite{van2018flexible}
    \item \textit{Regression based imputation}: We assume that the entries of the feature with missing values, is a function of the entries of the other features. Thus, we can formulate the problem of predicting the entries of the missing feature as a regression over the other features. Traditional machine learning algorithms like linear regression, logistic regression, support vector machines with different kernel settings can be used~\cite{van2018flexible}. A recent approach by~\cite{shrivastava2020grnular} modeled the regression as a multi-task learning problem and used deep neural networks to predict the missing values. This method is more robust than its traditional counterparts as it jointly optimizes for all the missing values.  
    \item \textit{Dropping the feature}: If the number of missing entries in the column is  significant, it is sometimes prudent to just drop the feature column as it might contribute more noise than information for the downstream tasks. The threshold to drop a column will most likely need to be decided on a case-by-case basis.
\end{itemize}

\textbf{Missing values in samples:}\label{sec:missing-data-samples}
It can often happen that there are technical errors or human mistakes in collecting samples, which can often lead to missing values or noise seeping into the sample. Also, we assume that the samples are independent and identically distributed (IID), so we cannot make use of the imputation techniques discussed in the case above. For this case, we propose a novel multi-task learning technique based on utilizing the \texttt{uGLAD}'s ability to optimize over a batch of input samples. 

\subsection{Consensus strategy: Multi-task learning over row-subsampled input}\label{sec:consensus-strategy}

The key idea is to create a batch of row subsamples of the input data $X\in \mathbb{R}^{M\times D}$. Since all of these subsamples come from the same underlying distribution, we should ideally recover the same precision matrix for the entire batch. Thus, if we have a model that can be jointly optimized over the entire batch for the \uglad objective, resulting in the recovered precision matrix being robust against erroneous or noisy samples. 



\begin{table}[]
    \centering
    \resizebox{0.60\columnwidth}{!}{
    \begin{tabular}{|c|c|c|c|}
    \hline
    \multicolumn{4}{|c|}{AUPR} \\
    \hline
    \hline
    Method & M=10            & M=25            & M=50           \\ \hline
    SGLCV & 0.112$\pm$0.013 & 0.132$\pm$0.012 & 0.219$\pm$0.085 \\ \hline
    \uglad  & 0.159$\pm$0.029 & 0.174$\pm$0.018 & 0.223$\pm$0.062 \\ \hline
    \end{tabular}
    }
    \resizebox{0.60\columnwidth}{!}{
    \begin{tabular}{|c|c|c|c|}
    \hline
    \multicolumn{4}{|c|}{AUC} \\
    \hline
    \hline
    Method & M=10            & M=25            & M=50           \\ \hline
    SGLCV & 0.505$\pm$0.007 & 0.532$\pm$0.031 & 0.617$\pm$0.083 \\ \hline
    \uglad  & 0.572$\pm$0.027 & 0.595$\pm$0.044 & 0.651$\pm$0.021 \\ \hline
    \end{tabular}
    }
    \vspace{1mm}
    \caption{\small\textbf{Synthetic data: Gaussian}. AUPR and AUC on $20$ test graphs for number of features $D=25$ and varying number of samples $M$. Gaussian Random graphs with sparsity $p=0.1$ were chosen and edge values were sampled from $\sim\mathcal{U}(-1, 1)$. We can observe that \uglad (CV mode) significantly outperforms SGLCV - the sklearn's graphicalLassoCV based on the BCD algorithm - for samples/features ratio << 1 and gives comparable performance as the number of samples increases.}
    \label{tab:compare-gaussian}
    \vspace{-3mm}
\end{table}

Steps for the multi-task learning approach to train the \uglad model for handling missing data:
\begin{enumerate}[leftmargin=*,nolistsep]
    \item \textit{Statistical imputation for the input}: Replace all the missing entries of the input data $X$ with their respective column mean (the mean is calculated ignoring the missing entries) $X[i,c]=mean(X[:,c])$. Replacing by mean is usually a preferred approach as its contribution zeros out $(X-\mu_X)$ while centering the data for the covariance matrix calculation.
    \item \textit{Getting the batches}: Perform stratified K-fold sampling to distribute the rows with missing values evenly among different batches. Say, we have $\mathbf{X_K} = [X_1, X_2, \cdots, X_K]$ batches with each $X_k\in\mathbb{R}^{M*(\frac{K-1}{K})\times D}$. Thus, the batch input for the \uglad model is $\mathbf{X_K}\in\mathbb{R}^{K\times M*(\frac{K-1}{K})\times D}$.
    \item \textit{Optimizing \texttt{uGLAD}}: It becomes a multi-task learning setting as we are jointly optimizing over a batch input $\mathbf{X_K}$. The \uglad model takes in the batch input $\mathbf{X_K}$ and outputs the corresponding $K$ precision matrices. Since, the entire batch of data is coming from the same underlying distribution, we use the entire data $X\in \mathbb{R}^{M\times D}$ for the \uglad loss to optimize the parameters of the \uglad model. Mathematically, we are minimizing the \uglad loss over the batch as
    \begin{align}
        \gL_\text{\ugladns-meta}&(\mathbf{X_K}) =\frac{1}{K}\sum_{k=1}^{K}\gL_\text{\ugladns}(\operatorname{cov}(X), f_{nn}(X_k))
    \end{align}
    \item \textit{Consensus among the batch to obtain the final precision matrix}: After optimizing the \uglad for the batch input, we will obtain $K$ different precision matrices $\Theta_K\in\mathbb{R}^{K\times D\times D}$. Ideally, all the precision matrices should be the same but there will be some discrepancies as we are working with missing values. Our `consensus' strategy to obtain the final precision matrix $\Theta^{f}$ is to find the common edges with their correlation type (positive or negative) from the batch precision matrices. Mathematically, we can obtain each entry $[i,j]$ of the final precision matrix as 
    \begin{align}
        \Theta_{i,j}^f = \operatorname{max-count}_{k=1,\ldots,K}(\operatorname{sign}\Theta_{i,j}^k) \min_{k=1,\ldots,K}|\Theta_{i,j}^k|
    \end{align}
    
    Here, the `max-count' term determines whether the correlation among the batches for that entry is positive or negative. The $2^{nd}$ term chooses the minimum absolute value for that entry among the batches as this ensures sparsity and conservative choice in terms of strength of an edge.  
\end{enumerate}

\begin{table}[]
    \centering
    \resizebox{0.6\columnwidth}{!}{
    \begin{tabular}{|c|c|c|c|}
    \hline
    \multicolumn{4}{|c|}{AUPR} \\
    \hline
    \hline
    Method & M=20            & M=100            & M=1000           \\ \hline
    SGLCV & 0.163$\pm$0.028 & 0.241$\pm$0.014 & 0.523$\pm$0.011 \\ \hline
    \uglad  & 0.206$\pm$0.035 & 0.272$\pm$0.024 & 0.569$\pm$0.048 \\ \hline
    \end{tabular}
    }
     \resizebox{0.6\columnwidth}{!}{
    \begin{tabular}{|c|c|c|c|}
    \hline
    \multicolumn{4}{|c|}{AUC} \\
    \hline
    \hline
    Method & M=20   & M=100     & M=1000           \\ \hline
    SGLCV & 0.670$\pm$0.013 & 0.718$\pm$0.014 & 0.839$\pm$0.006 \\ \hline
    \uglad  & 0.774$\pm$0.037 & 0.812$\pm$0.049 & 0.909$\pm$0.040 \\ \hline
    \end{tabular}
    }
        \vspace{1mm}
    \caption{\small\textbf{GRN data: non-Gaussian}. AUPR and AUC on $20$ test graphs for $D=100$ nodes and varying samples $M$. Graphs were sampled from the SERGIO simulator for the Gene Regulatory network recovery task. We can observe that the \uglad model is more adaptive in non-Gaussian settings. A post-hoc masking operation was done to remove all the edges not containing a transcription factor. This masking operation was done for all the methods.}
    \label{tab:compare-sergio-auc}
    \vspace{-2mm}
\end{table}

\begin{table}[]
    \centering
     \resizebox{0.6\columnwidth}{!}{
    \begin{tabular}{|c|c|c|c|}
    \hline
    \multicolumn{4}{|c|}{AUPR} \\
    \hline
    \hline
    Method & M=10  & M=25 & M=50            \\ \hline
    SGLCV-avg & 0.137$\pm$0.099 & 0.179$\pm$0.027 & 0.241$\pm$0.045 \\ \hline
    \texttt{uGLAD}-multi  &0.186$\pm$0.028 & 0.204$\pm$0.044 & 0.279$\pm$0.027\\ \hline
    \end{tabular}
    }
     \resizebox{0.6\columnwidth}{!}{
    \begin{tabular}{|c|c|c|c|}
    \hline
    \multicolumn{4}{|c|}{AUC} \\
    \hline
    \hline
    Method & M=10  & M=25 & M=50           \\ \hline
    SGLCV-avg & 0.508$\pm$0.024 & 0.538$\pm$0.024 & 0.597$\pm$0.047 \\ \hline
    \texttt{uGLAD}-multi  &0.552$\pm$0.048 & 0.573$\pm$0.047 & 0.626$\pm$0.022\\ \hline
    \end{tabular}
    }
    \vspace{1mm}
    \caption{\small\textbf{Multi-task learning}. Average AUPR and AUC over $K=10$ graphs coming from sparsity$\sim[0.05, 0.2]$. The number of nodes $D=25$ with varying samples $M=[10, 25, 50]$. The SGLCV-avg (the sklearn's graphicalLassoCV based on the BCD algorithm) considers each instance of the batch as a separate task and reports the average results over the batch. \texttt{uGLAD}-multi is used to recover the graphs jointly using a single model.}
    \label{tab:multi-task}
    \vspace{-2mm}
\end{table}

\section{Software details: Optimizing modes of \uglad}
To optimize \uglad for the \uglad loss function, we have introduced 4 different modes of training. In the software package, these modes can be switched from one to the other using an indicator flag.

\textit{Direct mode}: The input to the \uglad model is complete data $X$ and the output precision matrix $\Theta=f_{nn}(X)$ is optimized to reduce the \uglad loss $\gL_\text{\ugladns}(X)$, as defined in Eq.~\ref{eq:loss-uglad}. 

\textit{CV mode (recommended)}: In the k-fold cross validation mode, we split the input samples $X = (X_{train}, X_{valid})$. We use the $X_{train}$ as input and optimize for the \uglad loss $\gL_\text{\ugladns}(X_{\text{train}})$. Then, we select the best model that minimizes the $\gL_\text{\ugladns}(X_{\text{valid}})$ for the $X_{valid}$ samples. 


\textit{Missing data mode}: We give the entire data $X$ as input with a `NaN' indicator for the entries where the values are missing. The software then follows the `consensus' strategy for handling of missing data given in Sec.~\ref{sec:missing-data-samples} and outputs the final precision matrix $\Theta^{f}$. 

\textit{Multi-task mode}:  Given a batch of input data $X\in\mathbb{R}^{K\times M\times D}$, we jointly optimize them for the \uglad loss objective to obtain $K$ different precision matrices $\Theta_K\in\mathbb{R}^{K\times D\times D}$, refer Sec.~\ref{sec:multi-task}.

\begin{table}[]
    \centering
     \resizebox{0.6\columnwidth}{!}{
    \begin{tabular}{|c|c|c|c|}
    \hline
    \multicolumn{4}{|c|}{AUPR} \\
    \hline
    \hline
    Method & dp=0.25   & dp=0.50     & dp=0.75           \\ \hline
    SGLCV-mean & 0.583$\pm$0.082 & 0.335$\pm$0.012 & 0.100$\pm$0.009 \\ \hline
    \texttt{uGLAD}-mean  & 0.605$\pm$0.103 & 0.357$\pm$0.034 & 0.113$\pm$0.016\\ \hline
    \texttt{uGLAD}-missing  & 0.612$\pm$0.100 & 0.375$\pm$0.043 & 0.132$\pm$0.007 \\ \hline
    \end{tabular}
    }
     \resizebox{0.6\columnwidth}{!}{
    \begin{tabular}{|c|c|c|c|}
    \hline
    \multicolumn{4}{|c|}{AUC} \\
    \hline
    \hline
    Method & dp=0.25   & dp=0.50     & dp=0.75           \\ \hline
    SGLCV-mean & 0.792$\pm$0.045 & 0.649$\pm$0.005 & 0.508$\pm$0.009 \\ \hline
    \texttt{uGLAD}-mean  & 0.806$\pm$0.019 & 0.691$\pm$0.025 & 0.527$\pm$0.011 \\ \hline
    \texttt{uGLAD}-missing  & 0.815$\pm$0.010 & 0.718$\pm$0.002 & 0.560$\pm$0.0.41 \\ \hline
    \end{tabular}
    }
    \vspace{1mm}
    \caption{\small\textbf{Missing data: Gaussian}. AUPR and AUC on $20$ test graphs for $D=25$ nodes and samples $M=500$. Gaussian random graphs were generated as described in Sec.~\ref{sec:synthetic-gaussian}. Increasing fraction of dropouts were introduced to observe the robustness of handling missing data. We can observe that the \texttt{uGLAD}-missing model is more robust, especially in high dropout settings.}
    \label{tab:missing-gaussian}
    \vspace{-2mm}
\end{table}

\begin{table}[]
    \centering
     \resizebox{0.6\columnwidth}{!}{
    \begin{tabular}{|c|c|c|c|}
    \hline
    \multicolumn{4}{|c|}{AUPR} \\
    \hline
    \hline
    Method & dp=0.50   & dp=0.75     & dp=0.90           \\ \hline
    SGLCV-mean & 0.468$\pm$0.015 & 0.323$\pm$0.008 & 0.042$\pm$0.017 \\ \hline
    \texttt{uGLAD}-mean  & 0.503$\pm$0.011 & 0.346$\pm$0.021 & 0.090$\pm$0.069\\ \hline
    \texttt{uGLAD}-missing  & 0.523$\pm$0.004 & 0.361$\pm$0.043 & 0.117$\pm$0.093 \\ \hline
    \end{tabular}
    }
     \resizebox{0.6\columnwidth}{!}{
    \begin{tabular}{|c|c|c|c|}
    \hline
    \multicolumn{4}{|c|}{AUC} \\
    \hline
    \hline
    Method & dp=0.50   & dp=0.75     & dp=0.90           \\ \hline
    SGLCV-mean & 0.819$\pm$0.005 & 0.794$\pm$0.042 & 0.510$\pm$0.010 \\ \hline
    \texttt{uGLAD}-mean  & 0.897$\pm$0.009 & 0.821$\pm$0.019 & 0.598$\pm$0.079 \\ \hline
    \texttt{uGLAD}-missing  & 0.906$\pm$0.007 & 0.876$\pm$0.013 & 0.706$\pm$0.206 \\ \hline
    \end{tabular}
    }
    \vspace{1mm}
    \caption{\small\textbf{Missing data: GRN.} AUPR and AUC on $20$ test graphs for $D=100$ nodes and samples $M=1000$. Gene regulatory network data was used as described in Sec.~\ref{sec:grn-non-gaussian}. Increasing fraction of dropouts were introduced to the observed samples of the microarray expression data. \texttt{uGLAD}-missing model is more robust for high dropout settings. Such high dropout ratios are quite common in collecting samples for microarray gene expression data.}
    \label{tab:missing-grn}
    \vspace{-2mm}
\end{table}

\begin{figure}
    \centering
    \includegraphics[width=0.8\textwidth]{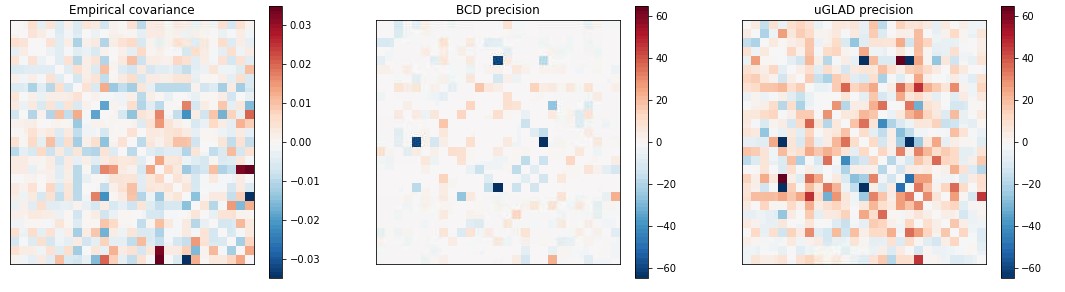} 
    \caption{\small  \uglad recovered precision matrix compared to empirical covariance and precision matrix recovered by the SGLCV algorithm for archaea at species level}
    \label{fig:archaea_species_covariance}
    \vspace{-3mm}
\end{figure}

\section{Experiments}\label{sec:expts}
Our software package is hosted on the GitHub
website. Its function signature is very much akin to the sklearn's GraphicalLassoCV~\cite{scikit-learn}. This was intended to make it easier for the users to try out our method with minimal change to their existing code pipeline. Sec.~\ref{apx:applications} lists some more potential applications of the \uglad model. We believe that \uglad can be seamlessly integrated with the existing pipeline for these applications and hope is that it improves results over state-of-the-art. 

We use AUC (area under the ROC curve) and AUPR (area under the precision-recall curve) as primary evaluation metrics. The sparsity of the graph leads to very few positive edges. These two metrics account for such imbalance in the data. In addition, they have the advantage of working without specifically setting a threshold for non-zero entries.  Their values reported in this work have the mean and the associated standard deviation values listed. 


\subsection{Performance on synthetically generated Gaussian samples}\label{sec:synthetic-gaussian}
The synthetic data was generated based on the procedure similar to the one described in \cite{guillot2012iterative}. A $d$-dimensional precision matrix $\Theta$ was generated by initializing a $d\times d$ matrix with its off-diagonal entries sampled i.i.d. from a uniform distribution $\Theta_{ij}\sim\mathcal{U}(-1, 1)$. These entries were then set to zero based on the sparsity pattern of the corresponding Erdos-Renyi random graph with a certain probability $p$. Finally, an appropriate multiple of the identity matrix was added to the current matrix, so that the resulting matrix had the smallest eigenvalue of $1$. In this way, $\Theta$ was ensured to be a well-conditioned, sparse and positive definite matrix and was used in the multivariate Gaussian distribution $\mathcal{N}(0, \Theta^{-1})$, to obtain $M$ i.i.d samples. Table~\ref{tab:compare-gaussian} shows the results on this synthetic data. 

\texttt{uGLAD} parameter settings: $\rho_{nn}$ was a 4 layer neural network and $\Lambda_{nn}$ was a 2 layer neural network. Both used 3 hidden units in each layer. The non-linearity used for hidden layers was $\tanh$, while the final layer had sigmoid ($\sigma$) as the non-linearity for both, $\rho_{nn}$ and $\Lambda_{nn}$ (refer Figure~\ref{fig:nn-design}). The learnable offset parameter of initial $\Theta_0$ was set to $t=1$. It was unrolled for $L=15$ iterations.The optimizer used was adam with the learning rates were chosen between $[0.0001, 0.001]$.

\begin{figure}
    \centering
    \includegraphics[width=0.8\textwidth]{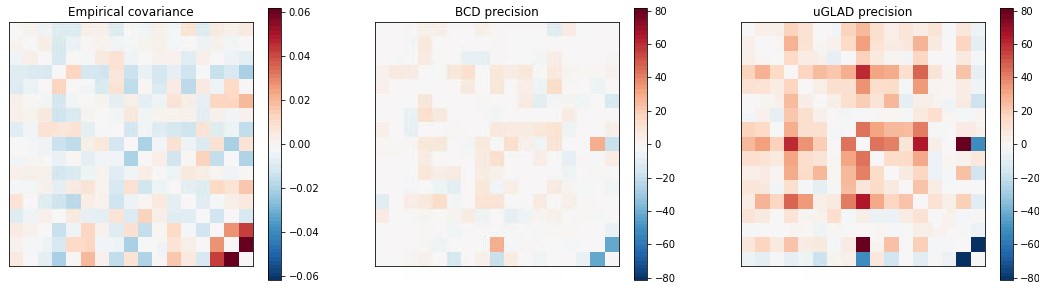}
    \caption{\small  \uglad recovered precision matrix compared to empirical covariance and precision matrix recovered by the SGLCV algorithm (sklearn's GraphicalLassoCV using BCD) for archaea at family level}
    \label{fig:archaea_family_covariance}
    \vspace{-3mm}
\end{figure}

\subsection{Recovery of Gene Regulatory Networks}\label{sec:grn-non-gaussian}
We conducted an exploratory study to gauge the generalization ability  of \uglad to non-Gaussian distributions. We chose the GRN inference task for this purpose. To this end, we use the SERGIO simulator~\cite{dibaeinia2020sergio} that provides a list of parameters to simulate cells from different types of biological processes and gene-expression levels with various amounts of intrinsic and technical noise.

For evaluation purposes in this work, we created random graphs (GRNs) that were used as input to SERGIO and will act as ground truth for evaluation. First, we set the number of Transcription Factors (TFs) or master regulators. Then, we randomly added edges between the TFs and the other genes based on sparsity requirements. Also, we randomly added some edges between the TFs themselves but excluded any self-regulation edges and maintained connectivity of the graph. 

When simulating data with no technical noise (clean data), we set the following parameters: sampling-state$=15$ (determines the number of steps of simulations for each steady state); noise-param $\sim U[0.1, 0.3]$ (controls the amount of intrinsic noise); noise-type = `dpd' (the type of intrinsic noise is dual production decay noise, which is the most complex out of all types provided); we set genes decay parameter to $1$.
The parameters required to decide the master regulators’ basal production cell rate for all cell types: low expression range of production cell rate $\sim U[0.2, 0.5]$ and high expression range of cell rate $\sim U[0.7, 1]$. We chose K$\sim U[1, 5]$, where K denotes the maximum interaction strength between master regulators and target genes. Positive strength values indicate activating interactions and negative strength values indicate repressive interactions and signs are randomly assigned. 
Table~\ref{tab:compare-sergio-auc} shows the comparison of the different graph recovery methods on the simulated data generated by SERGIO for cell types $C=5$ and clean data setting. 
\begin{figure}
    \centering
    \includegraphics[width=0.5\textwidth]{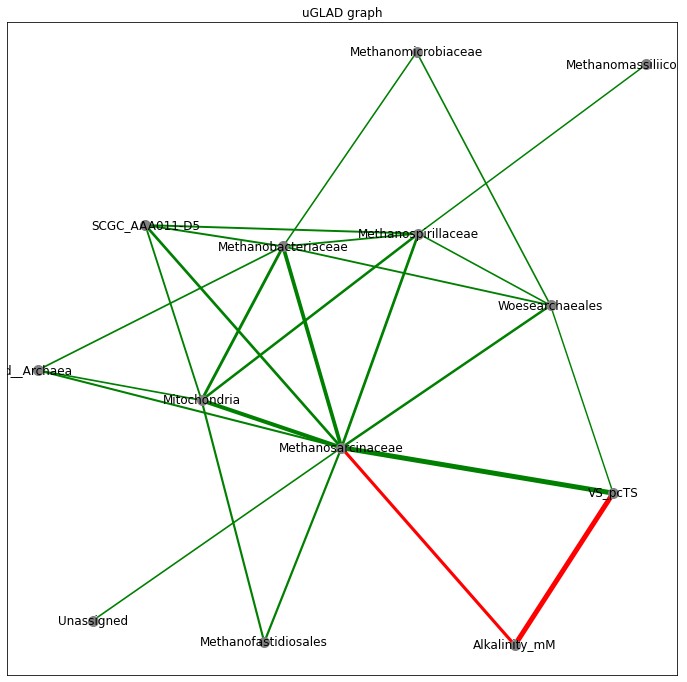} 
    \caption{\small  \uglad graph for archaea at family level. Edge color indicates the sign of the correlation: green - positive, red - negative, edge weight corresponds to correlation's strength.}
    \label{fig:archaea_family_graph}
    \vspace{-3mm}
\end{figure}

\begin{figure}
    \centering
    \includegraphics[width=0.5\textwidth]{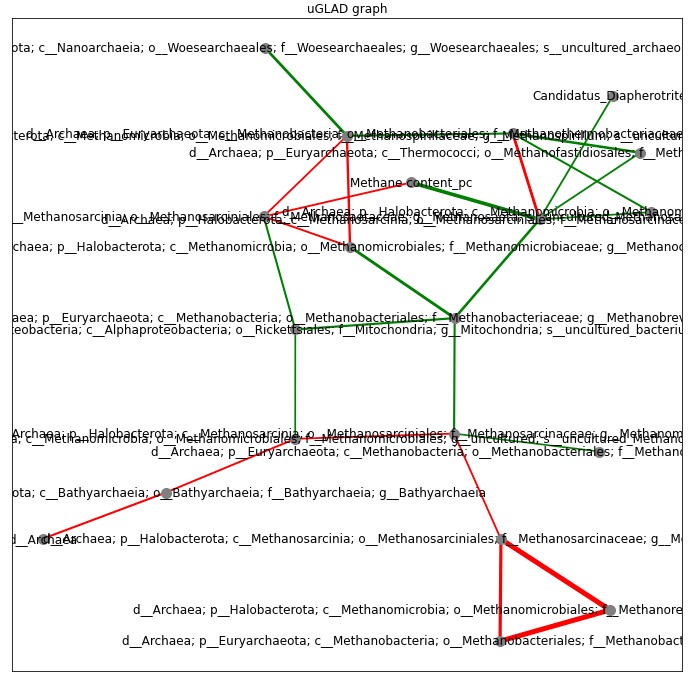} 
    \caption{\small \uglad graph for archaea at species level. Edge color indicates the sign of the correlation: green - positive, red - negative, edge weight corresponds to correlation's strength.}
    \label{fig:archaea_species_graph}
    \vspace{-3mm}
\end{figure}

\subsection{Multi-task learning experiments}
This experiment verifies the ability of \uglad to do multi-task learning. We chose a collection of tasks as a set of data coming from graphs with varying sparsity. For $K$ different tasks, our input data is $X\in\mathbb{R}^{K\times M\times D}$. We run the \uglad model in multi-task learning mode as described in Sec.~\ref{sec:multi-task} and recover $K$ different precision matrices $\hat{\Theta}\in\mathbb{R}^{K\times D\times D}$ that are optimized for the loss function $\gL_\text{\ugladns-multitask}(X_K)$ given by equation~\ref{eq:loss-mt-1}. Table~\ref{tab:multi-task} shows results from a single \uglad model in multi-task setting which is run on the synthetic data generated as described in Sec.~\ref{sec:synthetic-gaussian}. Our model ability to recover multiple graphs with varying sparsity shows promise for multitask learning. 


\subsection{Robustness testing for missing data}
We artificially introduced missing values or `dropouts' in the input data $X\in\mathbb{R}^{M\times D}$ to create noisy data. Our aim is to study the effectiveness of the `consensus' strategy discussed in Sec.~\ref{sec:consensus-strategy}. We compare it with the baseline statistical imputation technique that does column-wise (or feature-wise) mean imputation as a preprocessing step. SGLCV-mean and \texttt{uGLAD}-mean, report the results of running the corresponding methods on the column mean imputed data while the \texttt{uGLAD}-missing uses the `consensus' strategy. Tables~\ref{tab:missing-gaussian}\&\ref{tab:missing-grn} show the robustness of the `consensus' strategy introduced in this work for synthetic Gaussian data as well as for the Gene regulatory networks recovery tasks.

\begin{figure}
    \centering
    \includegraphics[width=0.5\textwidth]{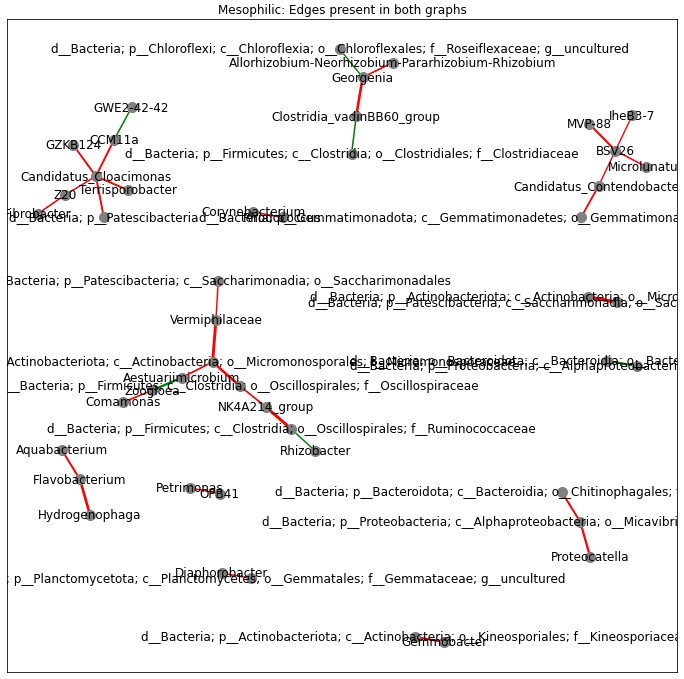}
    \caption{\small Example of multitask learning applied to different digester types.  \uglad graph for bacteria at genus level for mesophilic digesters showing only edges common to all digester types. Edge color indicates the sign of the correlation: green - positive, red - negative, edge weight corresponds to correlation's strength.}
    \label{fig:bacteria_genus_mesophilic}
\end{figure}

\section{Case study: Anaerobic Digestion}
Our algorithm development was inspired by a practical problem of domain exploration in anaerobic digestion.  Anaerobic digestion is a growing field addressing waste management with both environmental benefits (reduced odor and pathogens, improved soil health, reduction in methane emissions) and economic value from use of captured methane gas.  Despite numerous studies, the dynamics of organisms' growth in digesters, their dependence on conditions (temperature, pH, feedstock mix, nitrogen to carbon ratio, etc.) and their impact on methane yield are not well understood.  Our team's client, a large international company, requested an exploration tool for analyzing all these factors.

While we cannot present the results of our analysis on our client's proprietary data, we can share findings based on a public dataset from a study of anaerobic digesters at Danish wastewater plants \cite{JIANG2021116871}.  Data is available at NCBI under bioproject accession number PRJNA637463.

\begin{figure}
    \centering
    \includegraphics[width=0.5\textwidth]{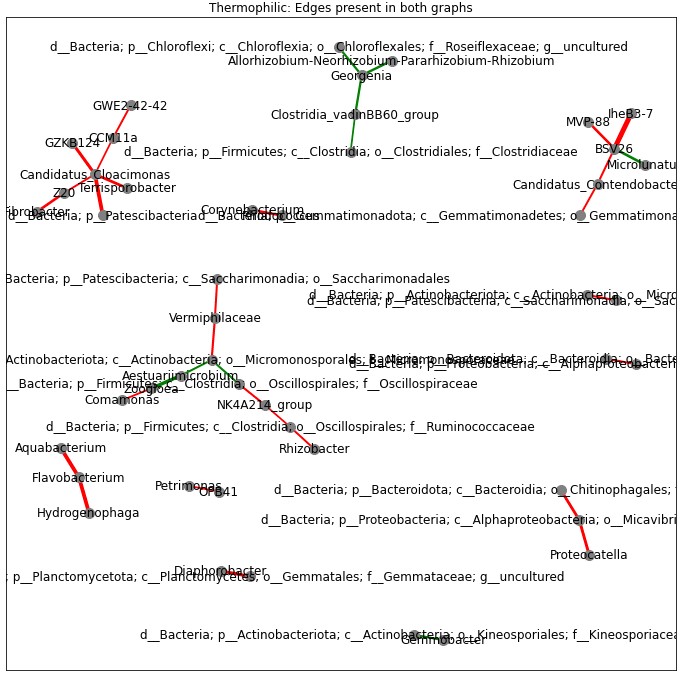} 
    \caption{\small Example of multitask learning applied to different digester types.  \uglad graph for bacteria at genus level for thermophilic digesters showing only edges common to all digester types. Note that edge colors (correlation signs) are different from the corresponding graph for mesophilic digesters. Edge color indicates the sign of the correlation: green - positive, red - negative, edge weight corresponds to correlation's strength.}
    \label{fig:bacteria_genus_thermophilic}
\end{figure}

Data comes from a 6-year study of 46 digesters located at 22 Danish treatment plants. We have three types of digesters, operating at different temperatures (mesophilic, mesophilic with thermal hydrolysis pretreatment, thermophilic). Digesters operate continuously with sludge retention rate of 15.8 to 35.6 days. Samples were taken at 3-month and 6-month intervals, so they can be treated as i.i.d.  We have a total of 1,010 sludge samples, 418 used to sequence archaea and 592 bacteria, performed using 16S rRNA gene amplicon sequencing. Analysis resulted in identification of 33,047 bacterial and 878 archaeal unique amplicon sequence variants (ASVs). 70\% of genera and 93\% of the species were determined to be novel or previously unclassified. This presents problems for all approaches attempting to utilize existing databases to determine organisms' function for the purpose of grouping and feature selection. In fact, one of the best ways to determine an organism's function is based on checking properties of organisms whose abundance numbers in the digester best correlate with the given organism's numbers.

Our algorithm works with any input, including: ASVs filtered by frequency, ASVs rolled up to higher taxonomy levels (species, genus, family), ASVs abundance normalized in various ways \cite{Badri406264}. We calculate the partial correlation matrix from the precision matrix. Each entry of the partial correlation matrix $\Rho_{ij}$ shows the correlation of the feature $x_i, x_j$ given the values of the other features are observed. This helps us obtain the direct dependence of the features. 
\begin{align}
\Rho_{ij} = -\frac{\Theta_{ij}}{\sqrt{\Theta_{ii}\Theta_{jj}}}
\end{align} 
We use networkx package to visualize the graphs, presenting positive correlations in green and negative in red, with edge weights corresponding to the strength of the correlation. Figures~\ref{fig:archaea_species_covariance} and  \ref{fig:archaea_family_covariance}  present precision matrices recovered by our algorithm and SGLCV with empirical covariance shown for comparison.  Figures~\ref{fig:archaea_family_graph} and \ref{fig:archaea_species_graph} show corresponding graphs for archaea at family and species level.

Figures~\ref{fig:bacteria_genus_mesophilic} and \ref{fig:bacteria_genus_thermophilic} show a result of multitask learning based on digester type: mesophilic (operating at temperature ~38\degree C) and thermophilic (operated at temperature ~53.6\degree C).  The two graphs' edges are filtered to show only edges common to both graphs, which is a small fraction of all edges.  Note that in some cases, the sign of the correlation (color of the edge) changes depending on digester's type.

Our model is being used by domain experts to gain insight into the domain of anaerobic digestion. One use case is to understand properties of newly discovered bacteria and archaea by analyzing which known organisms their abundance in digester samples correlates with (positively or negatively). That can lead to focusing attention on a smaller organism set. Another use case centers around understanding the role of digester conditions and feedstock mix on organisms' growth and methane yield. The results presented in recovered graphs lead to new hypotheses and new experiments being designed to test them.

The advantage of probabilistic graphical models over conventional predictive models is that they model the entire domain by fitting a probability distribution over all variables.  Thus they enable insight into the impact of any variable or a set of variables on any other variable in the domain, rather than focusing on the impact of all variables on a selected outcome variable.

\section{Potential applications of the \uglad model}\label{apx:applications}

Listing some more applications for the \uglad model for which we feel that it can help improve the current state-of-the-art performance.
\begin{itemize}[leftmargin=*,nolistsep]
    \item Protein Structure recovery: PSICOV~\cite{jones2012psicov} uses graphical lasso based approach to predict the contact matrix, which then eventually gives the 3D protein structure. \uglad can be substituted for predicting the contact matrix from the input correlation matrix between the amino acid sequences.
    \item Finance \& Healthcare: Finding correlations between stocks to see how companies compare~\cite{hallac2017network}. Similarly systems for finding connection between important body vitals of ICU patients~\cite{shrivastava2021system,bhattacharya2019methods}.
    \item Class imbalance handling: We can potentially use \uglad to find correlation between the features. This correlation graph can be helpful in sampling down useful feature clusters. This will help identify key features and in-turn improve performance in cases where there is less data or imbalanced data (more data points for one class over another). Some of the methods for class imbalance handling on which \uglad model can act as a preprocessing steps are~\cite{rahman2013addressing,shrivastava2015classification,bhattacharya2017icu}.
    \item Gaussian processes \& time series problems: \uglad can be extended to this interesting work by~\cite{chatrabgoun2021learning} on combining graphical lasso with Gaussian processes for learning gene regulatory networks. Similarly, in a recent work on including negative data points for the Gaussian processes~\cite{shrivastava2020learning}, \uglad can be used for narrowing down the relevant features for doing the GP regression. Our model can be used for time-series modeling, refer~\cite{jung2015graphical} for an example.  
    \item Video sequence predictions: \uglad can be integrated into the pipeline for latest models used for generating unseen future video frames~\cite{denton2018stochastic,shrivastava2021diverse}. Specifically, the parameters of the \uglad can be learned to narrow down the potential future viable frames from the generated ones.   
\end{itemize}

\section{Conclusions}
We introduce a novel technique~\uglad to perform sparse graph recovery by optimizing a deep unrolled network for the graphical lasso objective. This is an extension to the previous \texttt{GLAD} model that was designed to use supervision. The key advantages of using our model over the state-of-the-art algorithms for the graphical lasso problems are (1) Sparsity related hyperparameters are modeled using neural networks which are automatically learned during the optimization. We thus address the sensitivity issue of choosing the right sparsity parameters which is usually a tedious task and often manually set for the other algorithms. (2) By design, neural networks of \uglad enable the sparsity regularization to be adaptive over the iterations of the optimization leading to superior performance. (3) Our software implementation supports GPU based acceleration and thus can be scaled efficiently to meet the runtime requirements. (4) The \uglad framework can jointly learn over a batch of samples and can be used for doing multi-task learning. The proposed `consensus' strategy based on leveraging this property works well to robustly handle missing data. Our experiments on Gaussian synthetic data and non-Gaussian GRN data show very promising results.  For these reasons, we hope that our model becomes one the 
widely used algorithms to solve the graphical lasso objective. 

As our experience with anaerobic digestion demonstrates, sparse graph recovery can be successfully used as a tool for generating insight into growth dynamics of organisms in a digester and (hopefully) into domain structure in many other applications.

\bibliography{bibfile}

\begin{thebibliography}{53}
\providecommand{\natexlab}[1]{#1}
\providecommand{\url}[1]{\texttt{#1}}
\expandafter\ifx\csname urlstyle\endcsname\relax
  \providecommand{\doi}[1]{doi: #1}\else
  \providecommand{\doi}{doi: \begingroup \urlstyle{rm}\Url}\fi

\bibitem[Aluru et~al.(2021)Aluru, Shrivastava, Chockalingam, Shivakumar, and
  Aluru]{aluru2021engrain}
Maneesha Aluru, Harsh Shrivastava, Sriram~P Chockalingam, Shruti Shivakumar,
  and Srinivas Aluru.
\newblock {EnGRaiN}: a supervised ensemble learning method for recovery of
  large-scale gene regulatory networks.
\newblock \emph{Bioinformatics}, 2021.

\bibitem[Badri et~al.(2020)Badri, Kurtz, Bonneau, and M{\"u}ller]{Badri406264}
Michelle Badri, Zachary~D. Kurtz, Richard Bonneau, and Christian~L. M{\"u}ller.
\newblock Shrinkage improves estimation of microbial associations under
  different normalization methods.
\newblock \emph{bioRxiv}, 2020.
\newblock \doi{10.1101/406264}.
\newblock URL \url{https://www.biorxiv.org/content/early/2020/04/04/406264}.

\bibitem[Belilovsky et~al.(2017)Belilovsky, Kastner, Varoquaux, and
  Blaschko]{belilovsky2017learning}
Eugene Belilovsky, Kyle Kastner, Ga{\"e}l Varoquaux, and Matthew~B Blaschko.
\newblock Learning to discover sparse graphical models.
\newblock In \emph{International Conference on Machine Learning}, pages
  440--448. PMLR, 2017.

\bibitem[Bhattacharya et~al.(2017)Bhattacharya, Rajan, and
  Shrivastava]{bhattacharya2017icu}
Sakyajit Bhattacharya, Vaibhav Rajan, and Harsh Shrivastava.
\newblock {ICU} mortality prediction: a classification algorithm for imbalanced
  datasets.
\newblock In \emph{Proceedings of the AAAI Conference on Artificial
  Intelligence}, volume~31, 2017.

\bibitem[Bhattacharya et~al.(2019)Bhattacharya, Rajan, and
  Shrivastava]{bhattacharya2019methods}
Sakyajit Bhattacharya, Vaibhav Rajan, and Harsh Shrivastava.
\newblock Methods and systems for predicting mortality of a patient, November~5
  2019.
\newblock US Patent 10,463,312.

\bibitem[Cai et~al.(2011)Cai, Liu, and Luo]{cai2011constrained}
Tony Cai, Weidong Liu, and Xi~Luo.
\newblock A constrained $\ell_1$ minimization approach to sparse precision
  matrix estimation.
\newblock \emph{Journal of the American Statistical Association}, 106\penalty0
  (494):\penalty0 594--607, 2011.

\bibitem[Chatrabgoun et~al.(2021)Chatrabgoun, Soltanian, Mahjub, and
  Bahreini]{chatrabgoun2021learning}
H~Chatrabgoun, AR~Soltanian, H~Mahjub, and F~Bahreini.
\newblock Learning gene regulatory networks using {Gaussian} process emulator
  and graphical lasso.
\newblock \emph{Journal of Bioinformatics and Computational Biology}, page
  2150007, 2021.

\bibitem[Chen et~al.(2020)Chen, Li, Umarov, Gao, and Song]{chen2020rna}
Xinshi Chen, Yu~Li, Ramzan Umarov, Xin Gao, and Le~Song.
\newblock Rna secondary structure prediction by learning unrolled algorithms.
\newblock \emph{arXiv preprint arXiv:2002.05810}, 2020.

\bibitem[Danaher et~al.(2014)Danaher, Wang, and Witten]{danaher2014joint}
Patrick Danaher, Pei Wang, and Daniela~M Witten.
\newblock The joint graphical lasso for inverse covariance estimation across
  multiple classes.
\newblock \emph{Journal of the Royal Statistical Society. Series B, Statistical
  methodology}, 76\penalty0 (2):\penalty0 373, 2014.

\bibitem[Denton and Fergus(2018)]{denton2018stochastic}
Emily Denton and Rob Fergus.
\newblock Stochastic video generation with a learned prior.
\newblock In \emph{International Conference on Machine Learning}, pages
  1174--1183. PMLR, 2018.

\bibitem[Dibaeinia and Sinha(2020)]{dibaeinia2020sergio}
Payam Dibaeinia and Saurabh Sinha.
\newblock Sergio: a single-cell expression simulator guided by gene regulatory
  networks.
\newblock \emph{Cell Systems}, 11\penalty0 (3):\penalty0 252--271, 2020.

\bibitem[Friedman et~al.(2008)Friedman, Hastie, and
  Tibshirani]{friedman2008sparse}
Jerome Friedman, Trevor Hastie, and Robert Tibshirani.
\newblock Sparse inverse covariance estimation with the graphical lasso.
\newblock \emph{Biostatistics}, 9\penalty0 (3):\penalty0 432--441, 2008.

\bibitem[Gon{\c{c}}alves et~al.(2016)Gon{\c{c}}alves, Von~Zuben, and
  Banerjee]{gonccalves2016multi}
Andr{\'e}~R Gon{\c{c}}alves, Fernando~J Von~Zuben, and Arindam Banerjee.
\newblock Multi-task sparse structure learning with {Gaussian} copula models.
\newblock \emph{The Journal of Machine Learning Research}, 17\penalty0
  (1):\penalty0 1205--1234, 2016.

\bibitem[Guillot et~al.(2012)Guillot, Rajaratnam, Rolfs, Maleki, and
  Wong]{guillot2012iterative}
Dominique Guillot, Bala Rajaratnam, Benjamin~T Rolfs, Arian Maleki, and Ian
  Wong.
\newblock Iterative thresholding algorithm for sparse inverse covariance
  estimation.
\newblock \emph{arXiv preprint arXiv:1211.2532}, 2012.

\bibitem[Guo et~al.(2011)Guo, Levina, Michailidis, and Zhu]{guo2011joint}
Jian Guo, Elizaveta Levina, George Michailidis, and Ji~Zhu.
\newblock Joint estimation of multiple graphical models.
\newblock \emph{Biometrika}, 98\penalty0 (1):\penalty0 1--15, 2011.

\bibitem[Hallac et~al.(2017)Hallac, Park, Boyd, and
  Leskovec]{hallac2017network}
David Hallac, Youngsuk Park, Stephen Boyd, and Jure Leskovec.
\newblock Network inference via the time-varying graphical lasso.
\newblock In \emph{Proceedings of the 23rd ACM SIGKDD International Conference
  on Knowledge Discovery and Data Mining}, pages 205--213, 2017.

\bibitem[Heckerman(1995)]{heckerman1995a}
David Heckerman.
\newblock A tutorial on learning with {Bayesian} networks.
\newblock Technical Report MSR-TR-95-06, March 1995.
\newblock URL
  \url{https://www.microsoft.com/en-us/research/publication/a-tutorial-on-learning-with-bayesian-networks/}.

\bibitem[Heckerman and Nathwani(1992)]{heckerman1992toward2}
David Heckerman and Bharat~N. Nathwani.
\newblock Toward normative expert systems part {II}. {Probability-based}
  representations for efficient knowledge acquisition and inference.
\newblock \emph{Methods of Information in Medicine}, 31:\penalty0 106--116,
  August 1992.
\newblock URL
  \url{https://www.microsoft.com/en-us/research/publication/toward-normative-expert-systems-part-ii/}.

\bibitem[Heckerman et~al.(1992)Heckerman, Horvitz, and
  Nathwani]{heckerman1992toward1}
David Heckerman, Eric Horvitz, and Bharat~N. Nathwani.
\newblock Toward normative expert systems part {I}. {The} {Pathfinder} project.
\newblock \emph{Methods of Information in Medicine}, 31:\penalty0 90--105, June
  1992.
\newblock URL
  \url{https://www.microsoft.com/en-us/research/publication/toward-normative-expert-systems-part/}.

\bibitem[Honorio and Samaras(2010)]{honorio2010multi}
Jean Honorio and Dimitris Samaras.
\newblock Multi-task learning of {Gaussian} graphical models.
\newblock In \emph{ICML}, 2010.

\bibitem[Jiang et~al.(2021)Jiang, Peces, Andersen, Kucheryavskiy, Nierychlo,
  Yashiro, Andersen, Kirkegaard, Hao, Høgh, Hansen, Dueholm, and
  Nielsen]{JIANG2021116871}
Chenjing Jiang, Miriam Peces, Martin~Hjorth Andersen, Sergey Kucheryavskiy,
  Marta Nierychlo, Erika Yashiro, Kasper~Skytte Andersen, Rasmus~Hansen
  Kirkegaard, Liping Hao, Jan Høgh, Aviaja~Anna Hansen, Morten~Simonsen
  Dueholm, and Per~Halkjær Nielsen.
\newblock Characterizing the growing microorganisms at species level in 46
  anaerobic digesters at danish wastewater treatment plants: A six-year survey
  on microbial community structure and key drivers.
\newblock \emph{Water Research}, 193:\penalty0 116871, 2021.
\newblock ISSN 0043-1354.
\newblock \doi{https://doi.org/10.1016/j.watres.2021.116871}.
\newblock URL
  \url{https://www.sciencedirect.com/science/article/pii/S0043135421000695}.

\bibitem[Jones et~al.(2012)Jones, Buchan, Cozzetto, and
  Pontil]{jones2012psicov}
David~T Jones, Daniel~WA Buchan, Domenico Cozzetto, and Massimiliano Pontil.
\newblock {PSICOV}: precise structural contact prediction using sparse inverse
  covariance estimation on large multiple sequence alignments.
\newblock \emph{Bioinformatics}, 28\penalty0 (2):\penalty0 184--190, 2012.

\bibitem[Jung et~al.(2015)Jung, Hannak, and Goertz]{jung2015graphical}
Alexander Jung, Gabor Hannak, and Norbert Goertz.
\newblock Graphical lasso based model selection for time series.
\newblock \emph{IEEE Signal Processing Letters}, 22\penalty0 (10):\penalty0
  1781--1785, 2015.

\bibitem[Kolar et~al.(2010)Kolar, Song, Ahmed, Xing,
  et~al.]{kolar2010estimating}
Mladen Kolar, Le~Song, Amr Ahmed, Eric~P Xing, et~al.
\newblock Estimating time-varying networks.
\newblock \emph{Annals of Applied Statistics}, 4\penalty0 (1):\penalty0
  94--123, 2010.

\bibitem[Koller and Friedman(2009)]{Koller2009ProbabilisticGM}
Daphne Koller and Nir Friedman.
\newblock \emph{Probabilistic Graphical Models: Principles and Techniques}.
\newblock 2009.

\bibitem[Liu and Chen(2019)]{liu2019alista}
Jialin Liu and Xiaohan Chen.
\newblock {ALISTA}: Analytic weights are as good as learned weights in {LISTA}.
\newblock In \emph{International Conference on Learning Representations
  (ICLR)}, 2019.

\bibitem[Moerman et~al.(2019)Moerman, Aibar~Santos, Bravo Gonz{\'a}lez-Blas,
  Simm, Moreau, Aerts, and Aerts]{moerman2019grnboost2}
Thomas Moerman, Sara Aibar~Santos, Carmen Bravo Gonz{\'a}lez-Blas, Jaak Simm,
  Yves Moreau, Jan Aerts, and Stein Aerts.
\newblock {GRNBoost2 and Arboreto}: efficient and scalable inference of gene
  regulatory networks.
\newblock \emph{Bioinformatics}, 35\penalty0 (12):\penalty0 2159--2161, 2019.

\bibitem[Mohan et~al.(2014)Mohan, London, Fazel, Witten, and
  Lee]{mohan2014node}
Karthik Mohan, Palma London, Maryam Fazel, Daniela Witten, and Su-In Lee.
\newblock Node-based learning of multiple {Gaussian} graphical models.
\newblock \emph{The Journal of Machine Learning Research}, 15\penalty0
  (1):\penalty0 445--488, 2014.

\bibitem[Oyen and Lane(2012)]{oyen2012leveraging}
Diane Oyen and Terran Lane.
\newblock Leveraging domain knowledge in multitask {Bayesian} network structure
  learning.
\newblock In \emph{Proceedings of the AAAI Conference on Artificial
  Intelligence}, volume~26, 2012.

\bibitem[Pearl(1988)]{Pearl88}
Judea Pearl.
\newblock \emph{Probabilistic Reasoning in Intelligent Systems: Networks of
  Plausible Inference}.
\newblock Morgan Kaufmann, 1988.

\bibitem[Pedregosa et~al.(2011)Pedregosa, Varoquaux, Gramfort, Michel, Thirion,
  Grisel, Blondel, Prettenhofer, Weiss, Dubourg, Vanderplas, Passos,
  Cournapeau, Brucher, Perrot, and Duchesnay]{scikit-learn}
F.~Pedregosa, G.~Varoquaux, A.~Gramfort, V.~Michel, B.~Thirion, O.~Grisel,
  M.~Blondel, P.~Prettenhofer, R.~Weiss, V.~Dubourg, J.~Vanderplas, A.~Passos,
  D.~Cournapeau, M.~Brucher, M.~Perrot, and E.~Duchesnay.
\newblock Scikit-learn: Machine learning in {P}ython.
\newblock \emph{Journal of Machine Learning Research}, 12:\penalty0 2825--2830,
  2011.

\bibitem[Peterson et~al.(2015)Peterson, Stingo, and
  Vannucci]{peterson2015bayesian}
Christine Peterson, Francesco~C Stingo, and Marina Vannucci.
\newblock Bayesian inference of multiple {Gaussian} graphical models.
\newblock \emph{Journal of the American Statistical Association}, 110\penalty0
  (509):\penalty0 159--174, 2015.

\bibitem[Pratapa et~al.(2020)Pratapa, Jalihal, Law, Bharadwaj, and
  Murali]{pratapa2020benchmarking}
Aditya Pratapa, Amogh~P Jalihal, Jeffrey~N Law, Aditya Bharadwaj, and
  TM~Murali.
\newblock Benchmarking algorithms for gene regulatory network inference from
  single-cell transcriptomic data.
\newblock \emph{Nature methods}, 17\penalty0 (2):\penalty0 147--154, 2020.

\bibitem[Pu et~al.(2021)Pu, Cao, Zhang, Dong, and Chen]{pu2021learning}
Xingyue Pu, Tianyue Cao, Xiaoyun Zhang, Xiaowen Dong, and Siheng Chen.
\newblock Learning to learn graph topologies.
\newblock \emph{Advances in Neural Information Processing Systems}, 34, 2021.

\bibitem[Rahman and Davis(2013)]{rahman2013addressing}
M~Mostafizur Rahman and Darryl~N Davis.
\newblock Addressing the class imbalance problem in medical datasets.
\newblock \emph{International Journal of Machine Learning and Computing},
  3\penalty0 (2):\penalty0 224, 2013.

\bibitem[Ravikumar et~al.(2011)Ravikumar, Wainwright, Raskutti, and
  Yu]{ravikumar2011high}
Pradeep Ravikumar, Martin~J Wainwright, Garvesh Raskutti, and Bin Yu.
\newblock High-dimensional covariance estimation by minimizing $l_1$-penalized
  log-determinant divergence.
\newblock \emph{Electronic Journal of Statistics}, 5:\penalty0 935--980, 2011.

\bibitem[Shrivastava and Shrivastava(2021)]{shrivastava2021diverse}
Gaurav Shrivastava and Abhinav Shrivastava.
\newblock Diverse video generation using a {Gaussian} process trigger.
\newblock \emph{arXiv preprint arXiv:2107.04619}, 2021.

\bibitem[Shrivastava et~al.(2020{\natexlab{a}})Shrivastava, Shrivastava, and
  Shrivastava]{shrivastava2020learning}
Gaurav Shrivastava, Harsh Shrivastava, and Abhinav Shrivastava.
\newblock Learning what not to model: {Gaussian} process regression with
  negative constraints.
\newblock 2020{\natexlab{a}}.

\bibitem[Shrivastava(2020)]{shrivastava2020using}
Harsh Shrivastava.
\newblock \emph{On Using Inductive Biases for Designing Deep Learning
  Architectures}.
\newblock PhD thesis, Georgia Institute of Technology, 2020.

\bibitem[Shrivastava et~al.(2015)Shrivastava, Huddar, Bhattacharya, and
  Rajan]{shrivastava2015classification}
Harsh Shrivastava, Vijay Huddar, Sakyajit Bhattacharya, and Vaibhav Rajan.
\newblock Classification with imbalance: A similarity-based method for
  predicting respiratory failure.
\newblock In \emph{2015 IEEE international conference on bioinformatics and
  biomedicine (BIBM)}, pages 707--714. IEEE, 2015.

\bibitem[Shrivastava et~al.(2019)Shrivastava, Bart, Price, Dai, Dai, and
  Aluru]{shrivastava2019cooperative}
Harsh Shrivastava, Eugene Bart, Bob Price, Hanjun Dai, Bo~Dai, and Srinivas
  Aluru.
\newblock Cooperative neural networks ({CoNN}): Exploiting prior independence
  structure for improved classification.
\newblock \emph{arXiv preprint arXiv:1906.00291}, 2019.

\bibitem[Shrivastava et~al.(2020{\natexlab{b}})Shrivastava, Chen, Chen, Lan,
  Aluru, Liu, and Song]{shrivastava2020glad}
Harsh Shrivastava, Xinshi Chen, Binghong Chen, Guanghui Lan, Srinivas Aluru,
  Han Liu, and Le~Song.
\newblock {GLAD}: Learning sparse graph recovery.
\newblock In \emph{International Conference on Learning Representations},
  2020{\natexlab{b}}.
\newblock URL \url{https://openreview.net/forum?id=BkxpMTEtPB}.

\bibitem[Shrivastava et~al.(2020{\natexlab{c}})Shrivastava, Zhang, Aluru, and
  Song]{shrivastava2020grnular}
Harsh Shrivastava, Xiuwei Zhang, Srinivas Aluru, and Le~Song.
\newblock Grnular: Gene regulatory network reconstruction using unrolled
  algorithm from single cell rna-sequencing data.
\newblock \emph{bioRxiv}, 2020{\natexlab{c}}.

\bibitem[Shrivastava et~al.(2021)Shrivastava, Huddar, Bhattacharya, and
  Rajan]{shrivastava2021system}
Harsh Shrivastava, Vijay Huddar, Sakyajit Bhattacharya, and Vaibhav Rajan.
\newblock System and method for predicting health condition of a patient,
  August~10 2021.
\newblock US Patent 11,087,879.

\bibitem[Shrivastava et~al.(2022)Shrivastava, Zhang, Song, and
  Aluru]{shrivastava2022grnular}
Harsh Shrivastava, Xiuwei Zhang, Le~Song, and Srinivas Aluru.
\newblock {GRNUlar}: A deep learning framework for recovering single-cell gene
  regulatory networks.
\newblock \emph{Journal of Computational Biology}, 29\penalty0 (1):\penalty0
  27--44, 2022.

\bibitem[Song et~al.(2009)Song, Kolar, and Xing]{song2009keller}
Le~Song, Mladen Kolar, and Eric~P Xing.
\newblock Keller: estimating time-varying interactions between genes.
\newblock \emph{Bioinformatics}, 25\penalty0 (12):\penalty0 i128--i136, 2009.

\bibitem[Sun et~al.(2018)Sun, Tan, Liu, and Zhang]{sun2018graphical}
Qiang Sun, Kean~Ming Tan, Han Liu, and Tong Zhang.
\newblock Graphical nonconvex optimization via an adaptive convex relaxation.
\newblock In \emph{International Conference on Machine Learning}, pages
  4810--4817. PMLR, 2018.

\bibitem[Van~Buuren(2018)]{van2018flexible}
Stef Van~Buuren.
\newblock \emph{Flexible imputation of missing data}.
\newblock CRC press, 2018.

\bibitem[Varici et~al.(2021)Varici, Sihag, and Tajer]{varici2021learning}
Burak Varici, Saurabh Sihag, and Ali Tajer.
\newblock Learning shared subgraphs in ising model pairs.
\newblock In \emph{International Conference on Artificial Intelligence and
  Statistics}, pages 3952--3960. PMLR, 2021.

\bibitem[Witten et~al.(2011)Witten, Friedman, and Simon]{witten2011new}
Daniela~M Witten, Jerome~H Friedman, and Noah Simon.
\newblock New insights and faster computations for the graphical lasso.
\newblock \emph{Journal of Computational and Graphical Statistics}, 20\penalty0
  (4):\penalty0 892--900, 2011.

\bibitem[Yang et~al.(2015)Yang, Lu, Shen, Wonka, and Ye]{yang2015fused}
Sen Yang, Zhaosong Lu, Xiaotong Shen, Peter Wonka, and Jieping Ye.
\newblock Fused multiple graphical lasso.
\newblock \emph{SIAM Journal on Optimization}, 25\penalty0 (2):\penalty0
  916--943, 2015.

\bibitem[Yu et~al.(2019)Yu, Chen, Gao, and Yu]{yu2019dag}
Yue Yu, Jie Chen, Tian Gao, and Mo~Yu.
\newblock Dag-gnn: Dag structure learning with graph neural networks.
\newblock In \emph{International Conference on Machine Learning}, pages
  7154--7163. PMLR, 2019.

\bibitem[Zheng et~al.(2018)Zheng, Aragam, Ravikumar, and Xing]{zheng2018dags}
Xun Zheng, Bryon Aragam, Pradeep~K Ravikumar, and Eric~P Xing.
\newblock Dags with no tears: Continuous optimization for structure learning.
\newblock \emph{Advances in Neural Information Processing Systems},
  31:\penalty0 9472--9483, 2018.

\end{thebibliography}
\bibliographystyle{plainnat}

\end{document}